\documentclass[review]{elsarticle}

\usepackage{lineno,hyperref}
\usepackage{stmaryrd}
\usepackage{bm}
\usepackage{graphicx}
\usepackage{float}
\usepackage{graphicx}
\usepackage{stfloats}
\usepackage{amsmath}
\usepackage{epstopdf}
\usepackage{color}
\usepackage{multirow}

\journal{Neurocomputing}

\begin{document}

\begin{frontmatter}

\title{Group-based Sparse Representation for Image Compressive Sensing Reconstruction with Non-Convex Regularization}
\tnotetext[mytitlenote]{Fully documented templates are available in the elsarticle package on \href{http://www.ctan.org/tex-archive/macros/latex/contrib/elsarticle}{CTAN}.}

\author{Zhiyuan~Zha$^{a}$, Xinggan~Zhang$^{a}$, Qiong~Wang$^{a}$, Lan~Tang$^{a,b}$, Xin~Liu$^{c}$}
\address{$^{a}$ \footnotesize{School of Electronic Science and Engineering, Nanjing University, Nanjing 210023, China.} \\
    $^{b}$ \footnotesize{National Mobile Commun. Research Lab., Southeast University, Nanjing 210023, China.}
    $^{c}$ \footnotesize{The Center for Machine Vision and Signal Analysis, University of Oulu, 90014, Finland.}}

\begin{abstract}
Patch-based sparse representation modeling has shown great potential in image compressive sensing (CS) reconstruction. However, this model usually suffers from some limits, such as dictionary learning with great computational complexity, neglecting the relationship among similar patches. In this paper, a group-based sparse representation method with non-convex regularization (GSR-NCR) for image CS reconstruction is proposed. In GSR-NCR, the local sparsity and nonlocal self-similarity of images is simultaneously considered in a unified framework. Different from the previous methods based on sparsity-promoting convex regularization, we extend the non-convex weighted $\ell_p$ (0$<p<$1) penalty function on group sparse coefficients of the data matrix, rather than conventional $\ell_1$-based regularization. To reduce the computational complexity, instead of learning the dictionary with a high computational complexity from natural images, we learn the principle component analysis (PCA) based dictionary for each group. Moreover, to make the proposed scheme tractable and robust, we have developed an efficient iterative shrinkage/thresholding algorithm to solve the non-convex optimization problem. Experimental results demonstrate that the proposed method outperforms many state-of-the-art techniques for image CS reconstruction.
\end{abstract}


\end{frontmatter}


\section{Introduction}

Compressive sensing (CS) \cite{1,2,3}, which aims to recover signals from fewer measurements than suggested by the Nyquist sampling theory, is based on the hypothesis that the signals in question have compressible representations. The most attractive aspect of CS-based compression is that the sampling and compression are conducted simultaneously, and almost all computational cost is derived from the decoder stage, and thus, leading to a low computational cost of the encoder stage. Due to the superior property of CS, it has been widely applied to various areas, such as MRI image \cite{4}, remoting sensing \cite{5}, single-pixel camera \cite{6} and sensor networks \cite{7}.

In the theory of CS, if a signal is sparse in some transform domain, it is often sampled by the random projection and reconstructed by solving the $\ell_0$ minimization problem with the prior information which usually makes up the regularization terms. However, due to the fact that $\ell_0$ minimization is a difficult combinatorial optimization problem, solving this problem is NP-hard. For this reason, it has been proposed to replace the $\ell_0$ norm by its convex $\ell_1$ counterpart to make the optimization easy. For instance, Cand{\`e}s $\emph{et al}.$ \cite{1} proposed that solving $\ell_1$ minimization problem can recover a $K$-sparse signal $\textbf{\emph{X}}\in\Re^N$ from $M=O(Klog(N/K))$ random measurements. To solve the above $\ell_1$ minimization problem, many CS reconstruction algorithms have been proposed, such as linear programming \cite{3}, gradient projection sparse reconstruction \cite{8}, match pursuit \cite{9} and iterative thresholding \cite{10}.

As a basic image inverse problem in the filed of image restoration, maybe the hottest topic is image CS reconstruction, which has attracted a lot of research interest in the past few years \cite{20,21,11,12,22,24,25,26,13,14,15,16,23,17,50,18,42,19}. Image CS reconstruction aims to reconstruct high quality image from fewer measurements, which may even be far below the traditional Nyquist sampling rate. Due to the ill-posed nature of image CS reconstruction, it has been well-known that the prior knowledge of images plays a critical role in improving the performance of image CS reconstruction algorithms. Therefore, how to design an effective regularization term to describe the image priors is vital for image CS reconstruction tasks.

Early regularization models mainly consider the prior on the level of pixels, such as Tikhonov regularization \cite{27} and total variation (TV) regularization \cite{24,25,26}, utilize the local structure patterns of an image and high effectiveness to preserve image edges and recover the smooth regions. However, some undesirable properties are produced, including smearing out the image details and over-smoothing the images.

Another popular prior is based on image patch, which has shown promising performance in image CS reconstruction \cite{12,22,13,14}. The well-known work is sparse representation-based model \cite{28,29}, which assumes that image patch can be precisely encoded as a sparse linear combination of basic elements. These elements, called atoms, compose a dictionary \cite{30,31}.  The dictionary is usually learned from a natural image dataset \cite{51}.Compared with the traditional analytically designed dictionaries, such as DCT \cite{11,32} and wavelet \cite{16}, dictionaries learned directly from images are superior to be adapted to image local structures \cite{28,29}, and thus could improve the sparsity which results in better performance. For example, Zhang $\emph{et al}.$ \cite{13} proposed a method for image CS reconstruction using adaptively learned sparsifying basis via $\ell_0$ minimization. Zha $\emph{et al}.$ \cite{14} proposed an adaptive sparse nonlocal regularization (ASNR) model for image CS reconstruction. However, two main problems are still existing for patch-based sparse representation model. First, it is computationally expensive to learn an off-the-shelf dictionary. Second, this sparse representation-based model usually neglects the correlations between sparsely coded patches.

Image patches that have similar patterns can be spatially far from each other and thus can be collected in the whole image. The nonlocal self-similarity (NSS) prior characterizes the repetitiveness of textures and structures reflected by natural images within nonlocal regions, which can be exploited to retain  edges and  sharpness effectively. The seminal work of nonlocal means (NLM) denoising \cite{33} has motivated a wide range of studies on NSS and a flurry of NSS-based methods have been proposed for image CS reconstruction \cite{11,16,34}. For instance, Zhang $\emph{et al}.$ \cite{34} proposed a nonlocal total variation (NLTV) regularization model for image CS reconstruction. Zhang $\emph{et al}.$ \cite{11} proposed a framework via collaborative sparsity, which enforces local 2D sparsity and nonlocal 3D sparsity simultaneously, in an adaptive hybrid space-transform domain. Nasser $\emph{et al}.$ \cite{16} proposed a new technique for high-fidelity image CS reconstruction via joint adaptive sparsity regularization (JASR) in transform domain.

Recent advances have suggested that, by exploiting the NSS prior and clustering similar patches, group-based sparse representation has shown great potential in various image inverse problems \cite{35,36,37,38}. In this paper, we propose a new method for image CS reconstruction, using group-based sparse representation framework with non-convex regularization (GSR-NCR). The GSR offers a powerful mechanism of combining local sparsity and NSS of images simultaneously. Unlike the previous sparsity-promoting convex regularization methods, we extend the non-convex weighted $\ell_p$ (0$<p<$1) penalty function on group sparse coefficients of the data matrix, rather than conventional $\ell_1$-based regularization. In order to reduce the computational complexity, we learn the principle component analysis (PCA) based dictionary for each group to substitute for the dictionary with a high computational complexity learned from natural images. In addition, to make the optimization tractable, an efficient iterative shrinkage/thresholding algorithm is adopted to solve the non-convex optimization problem. Experimental results show that the proposed method can outperform many exisiting state-of-the-art image CS reconstruction methods.

The reminder of this paper is organized as follows. Section~\ref{2} briefly introduces CS theory, patch-based sparse representation modeling and group-based sparse representation modeling. Section~\ref{3} presents the modeling of group-based sparse representation with non-convex regularization (GSR-NCR) for image CS reconstruction and develops an iterative shrinkage/thresholding algorithm to solve the proposed GSR-NCR model. Section~\ref{4} presents the experimental results. Finally, some conclusions are given in Section~\ref{5}.

\section{Background and related work}
\label{2}
\subsection{Compressive Sensing}
Compressive sensing (CS) has attracted considerable attention from signal/image processing communities \cite{1,2,3}. In the theory of CS, $\textbf{\emph{X}}\in\Re^N$ is a finite length signal. $\textbf{\emph{X}}$ is said to be sparse if $\textbf{\emph{X}}$ can be represented  as a superposition of a small number of vectors taken from a known sparsifying transform domain basis $\boldsymbol\Psi$, such that $\boldsymbol\theta={\boldsymbol\Psi}^T\textbf{\emph{X}}$ contains only a small set of non-zero entries. The number of significant elements within the coefficient vector $\boldsymbol\theta$ is regarded as the quantitative criteria of the sparsity of $\textbf{\emph{X}}$ in $\boldsymbol\Psi$. To be concrete,  one seeks the perfect reconstruction of a signal $\textbf{\emph{X}}$ from its $M$ randomized linear measurements, i.e., $\textbf{\emph{Z}}={\boldsymbol\phi}\textbf{\emph{X}}$, where $\textbf{\emph{Z}}\in\Re^M$, ${\boldsymbol\phi}\in\Re^{M\times N}$ represents the random projection matrix and satisfies $M<N$. The goal of CS recovery is to reconstruct $\textbf{\emph{X}}$ from $\textbf{\emph{Z}}$ with subrate being $S=M/N$, which is usually formulated as the following $\ell_0$ minimization problem,
\begin{equation}
\arg\min_{{\boldsymbol\theta}} ||\boldsymbol\theta||_0, \qquad s.t. \qquad {\textbf{\emph{Z}}}={\boldsymbol\phi}{\boldsymbol\Psi}\boldsymbol\theta
\label{eq:1}
\end{equation} 
where $||*||_0$ is $\ell_0$-norm, counting the non-zero entries of ${\boldsymbol\theta}$.

However, since $||*||_0$ norm minimization is discontinuous and an NP-hard problem, it is usually relaxed to the convex $\ell_1$-norm minimization. Therefore, Eq.~\eqref{eq:1} can be rewritten as the following unconstrained optimization problem,
\begin{equation}
{\boldsymbol\theta}=\arg\min_{{\boldsymbol\theta}} \left(\frac{1}{2}||{\textbf{\emph{Z}}}-{\boldsymbol\phi}{\boldsymbol\Psi}\boldsymbol\theta||_2^2+\lambda||\boldsymbol\theta||_1\right)
\label{eq:2}
\end{equation} 
where $\lambda$ is regularization parameter. According to \cite{1}, CS is capable of recovering a $K$-sparse signal $\textbf{\emph{X}}$ (with  highly probability) from $\textbf{\emph{Z}}$ of size $M$, where the number of random measurements satisfies  $M=O(Klog(N/K))$.
\subsection{Patch-based Sparse Representation}
Traditional patch-based sparse representation model has been proven to be very effective in image CS reconstruction \cite{12,22,13,14}. It assumes that each image patch can be precisely modeled as a sparse linear combination of basic elements \cite{28,29}. These elements are called atoms and they compose a dictionary \cite{30,31}. Mathematically, for an image $\textbf{\emph{X}}\in\Re^N$, let $\textbf{\emph{x}}_i=\textbf{\emph{R}}_i\textbf{\emph{X}}$, $i=1,2,...n$ denotes an image patch of size $\sqrt{m} \times \sqrt{m}$ extracted at location $i$, where $\textbf{\emph{R}}_i$ is the matrix extracting patch $\textbf{\emph{x}}_i$ from $\textbf{\emph{X}}$ at location $i$. Given a dictionary $\textbf{\emph{D}}\in\Re^{m\times M}, m\leq M$, the sparse representation processing of each patch $\textbf{\emph{x}}_i$ is to discover a sparse vector ${\boldsymbol\alpha}_i$ such that ${\boldsymbol\alpha}_i=\textbf{\emph{D}}^{-1}\textbf{\emph{x}}_i$, where ${\boldsymbol\alpha}_i$ is a sparse vector whose entries are mostly zero or close to zero. Then the whole image $\textbf{\emph{X}}$ can be reconstructed by averaging all the reconstructed patches $\{\textbf{\emph{x}}_i\}$, which can be expressed as
\begin{equation}
\textbf{\emph{X}}\approx \textbf{\emph{D}}{\boldsymbol\alpha}=\left(\sum\limits_{i=1}^n{\textbf{\emph{R}}_i^T}{\textbf{\emph{R}}_i}\right)^{-1}
\left(\sum\limits_{i=1}^n{\textbf{\emph{R}}_i^T}{\textbf{\emph{D}}}{\boldsymbol\alpha_i}\right)
\label{eq:3}
\end{equation} 
where $\boldsymbol\alpha$ denotes the concatenation of all $\boldsymbol\alpha _i$, that is, $\boldsymbol\alpha=[\boldsymbol\alpha _1^T, \boldsymbol\alpha _2^T,...,\boldsymbol\alpha _n^T]^T$, which is the patch-based redundant sparse representation for $\textbf{\emph{X}}$.

Now, we merge Eq.~\eqref{eq:3} into Eq.~\eqref{eq:2}, the patch-based sparse representation scheme for image CS reconstruction is formulated as
\begin{equation}
{\boldsymbol\alpha}_i=\arg\min_{{\boldsymbol\alpha}_i} \sum_{i=1}^n\left(\frac{1}{2}||{\textbf{\emph{z}}}_i-{\boldsymbol\phi}{\textbf{\emph{D}}}\boldsymbol\alpha_i||_2^2+\lambda||\boldsymbol\alpha_i||_1\right)
\label{eq:4}
\end{equation} 
where $\textbf{\emph{D}}$ replaces $\boldsymbol\Psi$ in Eq.~\eqref{eq:2}, standing for a learning dictionary, and $\boldsymbol\alpha_i$ is a patch-based sparse representation coefficient for each patch ${\textbf{\emph{x}}}_i$ over the dictionary $\textbf{\emph{D}}$. ${\textbf{\emph{z}}}_i$ is the linear measurements of each patch ${\textbf{\emph{x}}}_i$.

However, there exists two main issues for patch-based sparse representation model. On one hand, since dictionary learning is a large-scale and highly non-convex problem, it is computationally expensive to solve the sparsity optimization problem. On the other hand, the patch-based sparse representation model usually assumes the independence between sparsely coded patches, which takes no account of  the correlation of similar patches in essence.
\begin{figure*}[!htbp]
\begin{minipage}[b]{1\linewidth}
  \centering
  \centerline{\includegraphics[width=12cm]{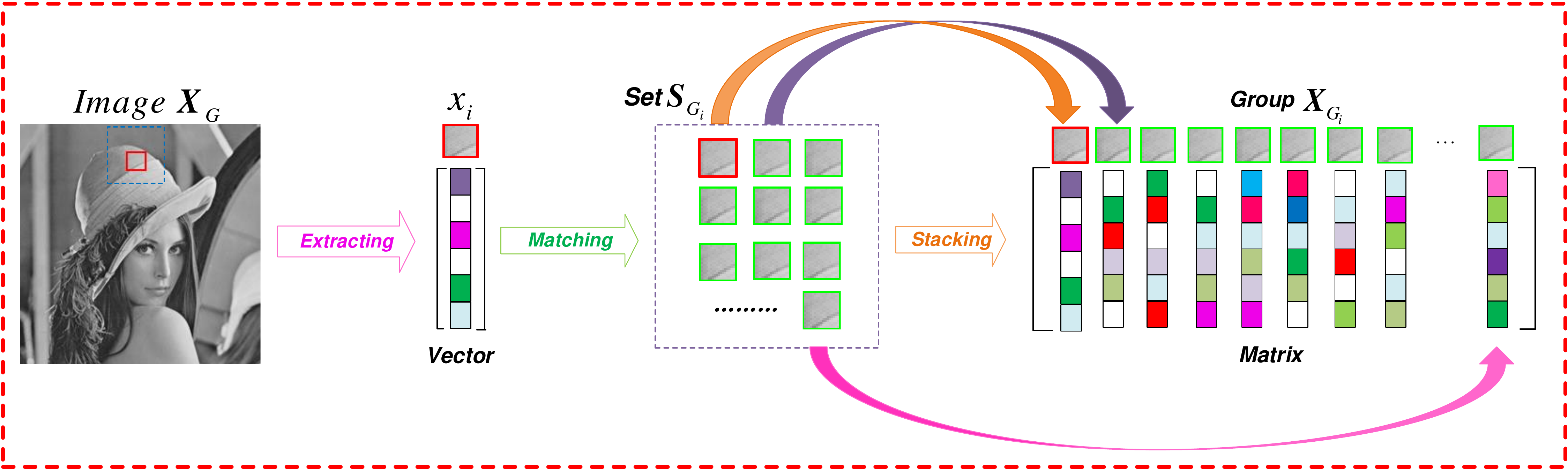}} 
\end{minipage}
\caption{Illustrations for the group construction. Extract each exemplar patch vector $\textbf{\emph{x}}_i$ from image $\textbf{\emph{X}}_G$. For each patch $\textbf{\emph{x}}_i$, denote ${\textbf{\emph{S}}}_{G_i}$ is the set, which composed of its most $c$ similar patches. Stack all the patches in ${\textbf{\emph{S}}}_{G_i}$ to construct the data matrix to generate the group, denoted by ${\textbf{\emph{X}}}_{G_i}$. }
\label{fig:1}
\end{figure*}
\subsection{Group-based Sparse Representation}
Recent studies have shown that structured or group sparsity can offer more promising performance for image restoration tasks \cite{35,36,37,38}. Since the unit of our proposed sparse representation model is group, this section will give briefs to introduce how to construct the groups. Specifically, as shown in Fig.~\ref{fig:1}, image  $\textbf{\emph{X}}_G$ with size $\emph{N}$ is divided into $\emph{n}$ overlapped patches $\textbf{\emph{x}}_i$ of size $\sqrt{m}\times\sqrt{m}, i=1,2,...,n$.  Then for each exemplar patch $\textbf{\emph{x}}_i$, denoted by small red square in Fig.~\ref{fig:1}, within the $H \times H$ sized searching window (big blue square), its most similar $c$ patches (small green squares) are selected to form a set ${\textbf{\emph{S}}}_{G_i}$. Since then, all the patches in ${\textbf{\emph{S}}}_{G_i}$ are stacked into a matrix ${\textbf{\emph{X}}}_{G_i}\in\Re^{{m}\times {c}}$, which contains every element of ${\textbf{\emph{S}}}_{G_i}$ as its column, i.e., ${\textbf{\emph{X}}}_{G_i}=\{{\textbf{\emph{x}}}_{G_{i,1}}, {\textbf{\emph{x}}}_{G_{i,2}}, ..., {\textbf{\emph{x}}}_{G_{i,c}}\}$.  The matrix ${\textbf{\emph{X}}}_{G_i}$ consisting of  all the patches with similar structures is called as a group, where ${\textbf{\emph{x}}_{G_{i,c}}}$ denotes the $c$-th similar patch (column form) of the $i$-th group. Finally, similar to patch-based sparse representation \cite{28,29}, given a dictionary ${\textbf{\emph{D}}}_{G_i}$, which is often learned from each group, such as DCT \cite{11}, PCA-based dictionary \cite{39}. Therefore, in image CS reconstruction,  similar to Eq.~\eqref{eq:4},  each group ${\textbf{\emph{X}}}_{G_i}$ can be sparsely represented as $\boldsymbol\alpha_{G_i}={{\textbf{\emph{D}}}_{G_i}}^{-1}\textbf{\emph{X}}_{G_i}$ and solved by the following $\ell_1$-norm minimization problem,
\begin{equation}
{\boldsymbol\alpha}_{G_i}=\arg\min_{{\boldsymbol\alpha}_{G_i}} \sum_{i=1}^n\left(\frac{1}{2}||{\textbf{\emph{Z}}}_{G_i}-{\boldsymbol\phi}{\textbf{\emph{D}}}_{G_i}\boldsymbol\alpha_{G_i}||_F^2+\lambda||\boldsymbol\alpha_{G_i}||_1\right)
\label{eq:5}
\end{equation} 
where ${\textbf{\emph{Z}}}_{G_i}$ is the linear measurements of each group ${\textbf{\emph{X}}}_{G_i}$.

\section{Image CS reconstruction using group-based sparse representation model with non-convex weighted $\ell_p$ Minimization}
\label{3}
Typical patch-based sparse representation methods for image CS reconstruction usually suffer from a common drawback that the dictionary learning with great computational complexity and neglecting the relationships among similar patches \cite{35,36,37,38}.  The sparsity-promoting convex $\ell_1$ minimization is usually regarded as a standard scheme for recovering a sparse signal. However, a fact that cannot be ignored is that, $\ell_1$ minimization is hard to achieve the desired sparsity solution in some practical problems, such as image inverse problems \cite{40}. Based on the fact above, this paper proposes a new method for image CS reconstruction using group-based sparse representation with non-convex weighted $\ell_p$ minimization. To make the optimization tractable, an  iterative shrinkage/thresholding (IST) algorithm  \cite{41} is developed to solve the above non-convex weighted $\ell_p$ minimization problem efficiently.

\subsection{Modeling of the Proposed Image CS Reconstruction}
To obtain sparsity solution more accurately, inspired by the success of $\ell_p$ ($0<p<1$) sparse optimization \cite{43,44,45} and our previous work \cite{40}, we apply the non-convex weighted $\ell_p$ ($0<p<1$) penalty function on group sparse coefficients of the data matrix to replace the convex $\ell_1$ norm. To be concrete, different from Eq.~\eqref{eq:5}, the proposed group-based sparse representation for image CS reconstruction with non-convex weighted $\ell_p$ minimization is formulated as
\begin{equation}
{\boldsymbol\alpha}_G=\arg\min_{{\boldsymbol\alpha}_G} \frac{1}{2}||{\textbf{\emph{Z}}}_G-{\boldsymbol\phi}{\textbf{\emph{D}}}_G\boldsymbol\alpha_G||_2^2+||{\textbf{\emph{W}}}_G\boldsymbol\cdot\alpha_G||_p
\label{eq:6}
\end{equation} 
where $\cdot$ represents the dot product and ${\textbf{\emph{W}}}_G$ is a weight assigned to $\boldsymbol\alpha_G$. The weight ${\textbf{\emph{W}}}_G$ will enhance the representation capability of group sparse coefficient $\boldsymbol\alpha_G$.
\subsection{Solving the Non-convex Weighted $\ell_p$ Minimization by the Iterative Shrinkage/Thresholding (IST) Algorithm}
 Solving the objective function of Eq.~\eqref{eq:6} is very difficult, since it is a large scale non-convex optimization problem. To make the proposed scheme tractable and robust, in this paper we adopt the iterative shrinkage/thresholding (IST) algorithm \cite{41} to solve Eq.~\eqref{eq:6}. We will briefly introduce IST algorithm. More specifically, consider the following general optimization problem,
 \begin{equation}
{ \rm min}_{\textbf{\emph{u}}\in\Re^N} f(\textbf{\emph{u}})+g(\textbf{\emph{u}})
\label{eq:7}
\end{equation} 
 where $f(\textbf{\emph{u}})$ is a smooth convex function with gradient,  which is Lipschitz continuous. $g(\textbf{\emph{u}})$ is a continuous convex function which is possibly non-smooth. The IST algorithm to solve Eq.~\eqref{eq:7} with a constant step $\rho$ is formulated as
\begin{equation}
\textbf{\emph{z}}^{(k+1)}=\textbf{\emph{u}}^{(k)}-\rho\nabla f(\textbf{\emph{u}}^{(k)})
\label{eq:8}
\end{equation} 
\begin{equation}
\textbf{\emph{u}}^{(k+1)}=\arg\min\limits_{\textbf{\emph{u}}}\frac{1}{2}||\textbf{\emph{u}}-\textbf{\emph{z}}^{(k+1)}||_2^2+\lambda g(\textbf{\emph{u}})
\label{eq:9}
\end{equation} 
where $k$ denotes the iteration number. Then, by invoking IST algorithm, the proposed non-convex weighted $\ell_p$ minimization problem Eq.~\eqref{eq:6} with the constraint $\textbf{\emph{u}}_G={\textbf{\emph{D}}}_G\boldsymbol\alpha_G$ can be rewritten as
\begin{equation}
{\textbf{\emph{u}}_G}^{(k)}={\textbf{\emph{D}}}_G^{(k)}\boldsymbol\alpha_G^{(k)}
\label{eq:10}
\end{equation} 
\begin{equation}
{\textbf{\emph{Y}}_G}^{(k+1)}={\textbf{\emph{u}}_G}^{(k)} -\rho {\boldsymbol\phi}^T(\boldsymbol\phi{\textbf{\emph{u}}_G}^{(k)}-{\textbf{\emph{Z}}}_G)
\label{eq:11}
\end{equation} 
\begin{equation}
\boldsymbol\alpha_G^{(k+1)}=\arg\min\limits_{\boldsymbol\alpha_G}\frac{1}{2}||{\textbf{\emph{D}}}_G\boldsymbol\alpha_G-{\textbf{\emph{Y}}}_G^{(k+1)}||_2^2+ ||{\textbf{\emph{W}}}_G\cdot\boldsymbol\alpha_G||_p
\label{eq:12}
\end{equation} 
Obviously, the crux for solving Eq.~\eqref{eq:6} is translated into solving Eq.~\eqref{eq:12}. Next, we will show that there is an efficient solution to Eq.~\eqref{eq:12}. To avoid confusion, the subscribe $k$ may be omitted for conciseness.

However, due to the complex structure of $||{\textbf{\emph{W}}}_G\boldsymbol\cdot\alpha_G||_p$, it is difficult to solve Eq.~\eqref{eq:12}, Let $\textbf{\emph{X}}_G={\textbf{\emph{D}}}_G\boldsymbol\alpha_G$, Eq.~\eqref{eq:12} can be rewritten as
\begin{equation}
\boldsymbol\alpha_G=\arg\min\limits_{\boldsymbol\alpha_G}\frac{1}{2}||\textbf{\emph{X}}_G-{\textbf{\emph{Y}}}_G||_2^2+ ||{\textbf{\emph{W}}}_G\cdot\boldsymbol\alpha_G||_p
\label{eq:13}
\end{equation} 

To enable a tractable solution of Eq.~\eqref{eq:13}, in this paper, a general assumption is made, with which even a closed-form solution can be achieved. Specifically, ${\textbf{\emph{Y}}}_G$ can be regarded as some type of noisy observation of $\textbf{\emph{X}}_G$, and then the assumption is made that each element of
$\textbf{\emph{E}}=\textbf{\emph{X}}_G-{\textbf{\emph{Y}}}_G$ follows an independent zero-mean distribution with variance ${
\sigma}^{2}$.  The following conclusion can be proved with this assumption.

\noindent$\textbf{Theorem 1}$ \ \ Define $\textbf{\emph{X}}_G,{\textbf{\emph{Y}}}_G\in\Re^{N}$, ${\textbf{\emph{X}}}_{G_i}$, ${\textbf{\emph{Y}}}_{G_i}\in\Re^{m\times c}$, and ${\textbf{\emph{e}}}{(j)}$ as each element of  error vector ${\textbf{\emph{e}}}$, where $\textbf{\emph{e}}=\textbf{\emph{X}}_G-\textbf{\emph{Y}}_G,  j=1,...,N$. Assume that ${\textbf{\emph{e}}}{(j)}$ follows an independent zero mean distribution with variance ${
\sigma}^{2}$, and thus for any $\varepsilon>0$, we can represent the relationship between $\frac{1}{N}||\textbf{\emph{X}}_G-\textbf{\emph{Y}}_G||_2^2$ and ${\frac{1}{K}}\sum_{i=1}^n||{\textbf{\emph{X}}}_{G_i}-{\textbf{\emph{Y}}}_{G_i}||_F^2$  by the following property,
\begin{equation}
\lim_{{N\rightarrow\infty}\atop{K\rightarrow\infty}}{\textbf{\emph{P}}}{\{|\frac{1}{N}||\textbf{\emph{X}}_G-\textbf{\emph{Y}}_G||_2^2
-{\frac{1}{K}}\sum\nolimits_{i=1}^n||{\textbf{\emph{X}}}_{G_i}-{\textbf{\emph{Y}}}_{G_i}||_F^2|<\varepsilon\}}=1
\label{eq:14}
\end{equation}
where ${\textbf{\emph{P}}}(\bullet)$  represents the probability and ${\emph{K}}=\emph{m}\times\emph{c}\times\emph{n}$. The detailed proof of $\emph{Theorem 1}$ can be seen in our previous work \cite{46}.

Therefore, based on $\emph{Theorem 1}$, we have the following equation with a very large probability (restricted 1) at each iteration,
\begin{equation}
\frac{1}{N}||\textbf{\emph{X}}_G-\textbf{\emph{Y}}_G||_2^2
={\frac{1}{K}}\sum\nolimits_{i=1}^n||{\textbf{\emph{X}}}_{G_i}-{\textbf{\emph{Y}}}_{G_i}||_F^2
\label{eq:15}
\end{equation}

Based on Eqs.~\eqref{eq:13} and~\eqref{eq:15}, we have
\begin{equation}
\begin{aligned}
&\min\limits_{{{\boldsymbol\alpha}}_G}\frac{1}{2}{||\textbf{\emph{X}}_G-\textbf{\emph{Y}}_G||_2^2}
 +||{\textbf{\emph{W}}}_G\cdot\boldsymbol\alpha_G||_p\\
&=\min\limits_{{{{\boldsymbol\alpha}}}_{G_i}}\sum\nolimits_{i=1}^n\left(\frac{1}{2}||{\textbf{\emph{X}}}_{G_i}-{\textbf{\emph{Y}}}_{G_i}||_F^2
 +\frac{K}{N}||{\textbf{\emph{W}}}_{G_i}\cdot\boldsymbol\alpha_{G_i}||_p\right)\\
 &=\min\limits_{{{{\boldsymbol\alpha}}}_{G_i}}\sum\nolimits_{i=1}^n\left(\frac{1}{2}||{\textbf{\emph{Y}}}_{G_i}-{{\textbf{\emph{D}}}_{G_i}{{{\boldsymbol\alpha}}}_{G_i}}||_F^2
 +\frac{K}{N}||{\textbf{\emph{W}}}_{G_i}\cdot\boldsymbol\alpha_{G_i}||_p\right)\\
  &=\min\limits_{{{{\boldsymbol\alpha}}}_{G_i}}\sum\nolimits_{i=1}^n\left(\frac{1}{2}||{\textbf{\emph{Y}}}_{G_i}-{{\textbf{\emph{D}}}_{G_i}{{{\boldsymbol\alpha}}}_{G_i}}||_F^2
 +\tau||{\textbf{\emph{W}}}_{G_i}\cdot\boldsymbol\alpha_{G_i}||_p\right)\\
\end{aligned}
\label{eq:16}
\end{equation}
where $\tau=K/N$.  Clearly, Eq.~\eqref{eq:16} can be regarded as a sparse representation problem by solving $n$ sub-problems for all the group ${\textbf{\emph{X}}}_{G_i}$.

Note that, dictionary learning are often learned from images, but we have only the linear measurements ${{\textbf{\emph{Z}}}_{G}}$. Thus, we need to generate a initial image from the linear measurements ${{\textbf{\emph{Z}}}_{G}}$. In this paper, we first use the Multi-hypothesis block-based compressive sensing (MH-BCS) method \cite{50} to generate the initial image ${{\textbf{\emph{X}}}_{G}}$. To adapt to the local image structures, instead of learning an over-complete dictionary for each group as in \cite{35}, we learn the principle component analysis (PCA) based dictionary \cite{39} for each group ${\textbf{\emph{Y}}}_{G_i}$. The  proposed PCA-based dictionary learning method is efficient and convenient since it only requires one PCA decomposition operator for each group ${\textbf{\emph{Y}}}_{G_i}$, rather than learning the dictionary from natural image dataset with a high computational complexity.

Due to the fact that each dictionary ${{\textbf{\emph{D}}}_{G_i}}$ is orthogonal, Eq.~\eqref{eq:16} is equal to the following formula:
\begin{equation}
\begin{aligned}
&{\hat{\boldsymbol\alpha}}_{G_i}=\min\limits_{{{{\boldsymbol\alpha}}}_{G_i}}\sum\nolimits_{i=1}^n\left(\frac{1}{2}||{{{{\boldsymbol\gamma}}}_{G_i}}-{{{{\boldsymbol\alpha}}}_{G_i}}||_F^2
 +\tau||{\textbf{\emph{W}}}_{G_i}\cdot\boldsymbol\alpha_{G_i}||_p\right)\\
&=\min\limits_{\tilde{{{\boldsymbol\alpha}}}_{G_i}}\sum\nolimits_{i=1}^n\left(\frac{1}{2}||{\tilde{{{\boldsymbol\gamma}}}_{G_i}}-{\tilde{{{\boldsymbol\alpha}}}_{G_i}}||_2^2
 +\tau||{\tilde{\textbf{\emph{w}}}}_{G_i}\cdot \tilde{\boldsymbol\alpha}_{G_i}||_p\right)\\
 \end{aligned}
\label{eq:17}
\end{equation}
where ${\textbf{\emph{Y}}}_i={{\textbf{\emph{D}}}_{G_i}{{{\boldsymbol\gamma}}}_{G_i}}$ and
${\textbf{\emph{X}}}_i={{\textbf{\emph{D}}}_{G_i}{{{\boldsymbol\alpha}}}_{G_i}}$. ${\tilde{{{\boldsymbol\alpha}}}_{G_i}}$, ${\tilde{{{\boldsymbol\gamma}}}_{G_i}}$ and ${\tilde{\textbf{\emph{w}}}}_{G_i}$ denote the vectorization of the matrix ${{{{\boldsymbol\alpha}}}_{G_i}}$, ${{{{\boldsymbol\alpha}}}_{G_i}}$ and ${\textbf{\emph{W}}}_{G_i}$, respectively.

To obtain the solution of Eq.~\eqref{eq:17} effectively, in this paper, the generalized soft-thresholding (GST) algorithm \cite{45} is adopted to solve Eq.~\eqref{eq:17}. Specifically, given $p$, ${\tilde{{{\boldsymbol\gamma}}}_{G_i}}$ and ${\tilde{\textbf{\emph{w}}}}_{G_i}$, there exists a specific threshold,
\begin{equation}
\tau_p^{\emph{GST}}({\tilde{{\emph{w}}}_{G_{i,j}}})=(2{\tilde{{\emph{w}}}_{G_{i,j}}}(1-p))^{\frac{1}{2-p}}+{\tilde{{\emph{w}}}_{G_{i,j}}}p(2{\tilde{{\emph{w}}}_{G_{i,j}}}(1-p))^{\frac{p-1}{2-p}}
\label{eq:18}
\end{equation} 
where ${\tilde{\gamma}_{G_{i,j}}}$, ${\tilde{\alpha}_{G_{i,j}}}$ and ${\tilde{{\emph{w}}}_{G_{i,j}}}$ are the $j$-th element of  ${\tilde{{{\boldsymbol\gamma}}}_i}$, ${\tilde{{{\boldsymbol\alpha}}}_i}$ and  ${\tilde{\textbf{\emph{w}}}}_i$, respectively. Then, if ${\tilde{\gamma}_{G_{i,j}}}<\tau_p^{\emph{GST}}({\tilde{{\emph{w}}}_{G_{i,j}}})$, ${\tilde{\alpha}_{G_{i,j}}}=0$ is the global minimum. Otherwise, the optimum will be achieved at non-zero point. According to \cite{45}, for any ${\tilde{\gamma}_{G_{i,j}}}\in(\tau_p^{\emph{GST}}({\tilde{{\emph{w}}}_{G_{i,j}}}), +\infty)$, Eq.~\eqref{eq:17} has one unique minimum ${\textbf{\emph{S}}}_p^{\emph{GST}}({\tilde{\gamma}_{G_{i,j}}}; {\tilde{{\emph{w}}}_{G_{i,j}}})$, which can be obtained by solving the following equation,
\begin{equation}
{\textbf{\emph{S}}}_p^{\emph{GST}}({\tilde{\gamma}_{G_{i,j}}}; {\tilde{{\emph{w}}}_{G_{i,j}}})- {\tilde{\gamma}_{G_{i,j}}} + {\tilde{{\emph{w}}}_{G_{i,j}}}p \left({\textbf{\emph{S}}}_p^{\emph{GST}}({\tilde{\gamma}_{G_{i,j}}}; {\tilde{{\emph{w}}}_{G_{i,j}}})\right)^{p-1} =0
\label{eq:19}
\end{equation} 

The complete description of the GST algorithm is shown in Algorithm 1.  For more details about the GST algorithm, please refer to \cite{45}.
\begin{table}[!htbp]
\centering  
\begin{tabular}{lccc}  
\hline  
\qquad \ \  Algorithm 1: Generalized Soft-Thresholding (GST) \cite{45}.\\
\hline
$\textbf{Input:}$ \ ${\tilde{\gamma}_{G_{i,j}}}, {\tilde{{\emph{w}}}_{G_{i,j}}}, p, J$.\\
\hline
1.\ \ \ $\tau_p^{\emph{GST}}({\tilde{{\emph{w}}}_{G_{i,j}}})=(2{\tilde{{\emph{w}}}_{G_{i,j}}}(1-p))^{\frac{1}{2-p}}+{\tilde{{\emph{w}}}_{G_{i,j}}}p(2{\tilde{{\emph{w}}}_{G_{i,j}}}(1-p))^{\frac{p-1}{2-p}}
$;\\
2.\ \ \ $\textbf{If}$ \ \ \ $|{\tilde{\gamma}_{G_{i,j}}}|\leq \tau_p^{\emph{GST}}({\tilde{{\emph{w}}}_{G_{i,j}}})$\\
3.\ \ \ \ \ \ \ \ \ ${\textbf{\emph{S}}}_p^{\emph{GST}}({\tilde{\gamma}_{G_{i,j}}}; {\tilde{{\emph{w}}}_{G_{i,j}}})=0$;\\
4.\ \ \ $\textbf{else}$\\
5.\ \ \ \ \ \ \ \ \ $k=0, {\tilde{\alpha}_{G_{i,j}}}^{(k)}=|{\tilde{\gamma}_{G_{i,j}}}|$;\\
6.\ \ \ \ \ \ \ \ \ Iterate on $k=0, 1, ..., J$\\
7.\ \ \ \ \ \ \ \ \ ${\tilde{\alpha}_{G_{i,j}}}^{(k+1)}=|{\tilde{\gamma}_{G_{i,j}}}|-{{\tilde{\emph{w}}}_{G_{i,j}}}p \left({\tilde{\alpha}_{G_{i,j}}}^{(k)}\right)^{p-1}$;\\
8.\ \ \ \ \ \ \ \ \ $k\leftarrow k+1$;\\
9.\ \ \ \ \ \ \ \ \ ${\textbf{\emph{S}}}_p^{\emph{GST}}({\tilde{\gamma}_{G_{i,j}}}; {\tilde{{\emph{w}}}_{G_{i,j}}})={\rm sgn}({\tilde{\gamma}_{G_{i,j}}}){\tilde{\alpha}_{G_{i,j}}}^{k}$;\\
10.\ \  $\textbf{End}$\\
$\textbf{Ouput:}$ ${\textbf{\emph{S}}}_p^{\emph{GST}}({\tilde{\gamma}_{G_{i,j}}}; {\tilde{{\emph{w}}}_{G_{i,j}}})$.\\
\hline
\end{tabular}
\end{table}

Each weight $\textbf{\emph{W}}_{G_i}$ is assigned to group sparse coefficient $\boldsymbol\alpha_{G_i}$, large values of each $\boldsymbol\alpha_{G_i}$ usually include major edge and texture information. This implies that to reconstruct ${\textbf{\emph{X}}}_{G_i}$ from its degraded one, we should shrink large values less, while shrinking smaller ones more \cite{47}. Inspired by \cite{48}, the weight $\textbf{\emph{W}}_{G_i}$ of each group  is set as ${\tilde{\textbf{\emph{w}}}}_{G_{i}}= [{\tilde{{\emph{w}}}}_{G_{i,1}}, {\tilde{{\emph{w}}}}_{G_{i,2}}, ..., {\tilde{{\emph{w}}}}_{G_{i,j}}]$ and we have
\begin{equation}
{\tilde{{\emph{w}}}}_{G_{i,j}}=\frac{2\sqrt{2}\sigma^2}{(\tilde{\boldsymbol\delta}_{G_i}+\epsilon)}
\label{eq:20}
\end{equation} 
where $\tilde{\boldsymbol\delta}_{G_i}$ denotes the estimated variance of each group sparse coefficient $\tilde{{{{\boldsymbol\gamma}}}}_{G_i}$, and $\epsilon$ is a small constant.  Obviously, it can be seen that  each value of weight $\textbf{\emph{W}}_{G_i}$ is inverse proportion to each value of ${{{\boldsymbol\gamma}}}_{G_i}$ \cite{47}.
In light of all derivations, the complete description of the proposed image CS reconstruction using group-based sparse representation via non-convex weighted $\ell_p$ minimization is given in Table~\ref{lab:1}.
\begin{table}[!htbp]
\caption{The proposed GSR-NCR method for Image CS reconstruction.}
\centering  
\begin{tabular}{lccc}  
\hline  
$\textbf{Input:}$ \ The observed measurement $\textbf{\emph{Y}}_G$, the measurement matrix $\boldsymbol\phi$.\\
  $\rm \textbf{Initialization:} $\  Estimate an initial image ${\textbf{\emph{X}}_G}^{(0)}$ using a MH-BCS method \cite{50} and \\
  \qquad \qquad \qquad set parameters  \emph{m}, \emph{c}, $\rho$, \emph{p}, $\sigma$, $\epsilon$, \emph{H}, \emph{J};\\
  $\rm \textbf{For}$\ $k=1, 2, ..., Max\_iter$ $\rm \textbf{do}$\\
  \qquad \qquad Update $\textbf{\emph{Y}}_{G}^{k+1}$ computing by Eq.~\eqref{eq:11}.\\
   \qquad \qquad Generating the groups $\textbf{\emph{Y}}_{G_i}$ by searching similar patches from $\textbf{\emph{Y}}_{G}$.\\
 \qquad $\rm \textbf{For}$\ each group $\textbf{\emph{Y}}_{G_i}$ $\rm \textbf{do}$\\
   \qquad \qquad Constructing dictionary ${{{\textbf{\emph{D}}}}_{G_i}}^{k+1}$ for ${{{\textbf{\emph{Y}}}}_{G_i}}$ by PCA operator.\\
  \qquad \qquad Update ${{{\boldsymbol\gamma}}}_{G_i}^{k+1}$ by ${{{\boldsymbol\gamma}}}_{G_i}={{{\textbf{\emph{D}}}}_{G_i}}^{-1}{{{\textbf{\emph{Z}}}}_{G_i}}$.\\
   \qquad \qquad Update ${\textbf{\emph{W}}_{G_i}}^{k+1}$  by Eq.~\eqref{eq:20}.\\
   \qquad \qquad Update $\boldsymbol\alpha_{G_i}^{k+1}$ computing by Algorithm 1.\\
   \qquad \qquad Get the estimation ${{{\textbf{\emph{X}}}}_{G_i}}^{k+1}$ =${{{\textbf{\emph{D}}}}_{G_i}}^{k+1}$$\boldsymbol\alpha_{G_i}^{k+1}$.\\
  \qquad  $\rm \textbf{End for}$\\
   \qquad \qquad Aggregate all group ${{{\textbf{\emph{X}}}}_{G_i}}^{k+1}$ to form the recovered image ${{\hat{\textbf{\emph{X}}}}_G}^{k+1}$.\\
    $\rm \textbf{End for}$\\
     $\textbf{Output:}$ ${{\hat{\textbf{\emph{X}}}}_G}^{k+1}$.\\
\hline
\end{tabular}
\label{lab:1}
\end{table}
\begin{figure*}[!htbp]
\begin{minipage}[b]{1\linewidth}
  \centering
  \centerline{\includegraphics[width=12cm]{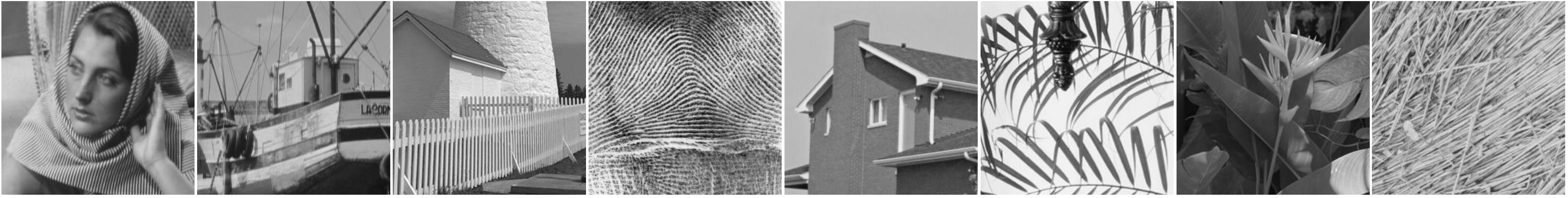}} 
\end{minipage}
\caption{All test images. From left to right: Barbara, boats, Fence, F.print, House, Leaves, plants, staw.}
\label{fig:2}
\end{figure*}

\section{Experimental Results}
\label{4}
In this section, we will report the experimental results of the proposed GSR-NCR for image CS reconstruction. All the experimental images are shown in Fig.~\ref{fig:2}. To evaluate the quality of the restored images, the PSNR and the recently proposed powerful perceptual quality metric FSIM \cite{49} are calculated.  
\subsection {Parameter Setting}
We generate the CS measurements at the block level by using a Gaussian random projection matrix to test images, i.e., the block-based CS reconstruction with block size of $32\times 32$. The parameters are set as follows. The size of each patch $\sqrt{m}\times \sqrt{m}$ is set to be $7\times 7$. Similar patch numbers $c=60$, the search window size $H=20$, $\sigma=\sqrt{2}$, $\epsilon=10^{-14}$, $J=2$. ($\rho, p$) are set to (0.3, 0.5), (1.5, 0.95) and (1.5, 0.95) when $0.2N$, $0.3N$ and $0.4N$, respectively.

\subsection {Performance Comparison with the State-of-the-Art methods}
\begin{figure*}[!htbp]
\begin{minipage}[b]{1\linewidth}
  \centering
  \centerline{\includegraphics[width=12cm]{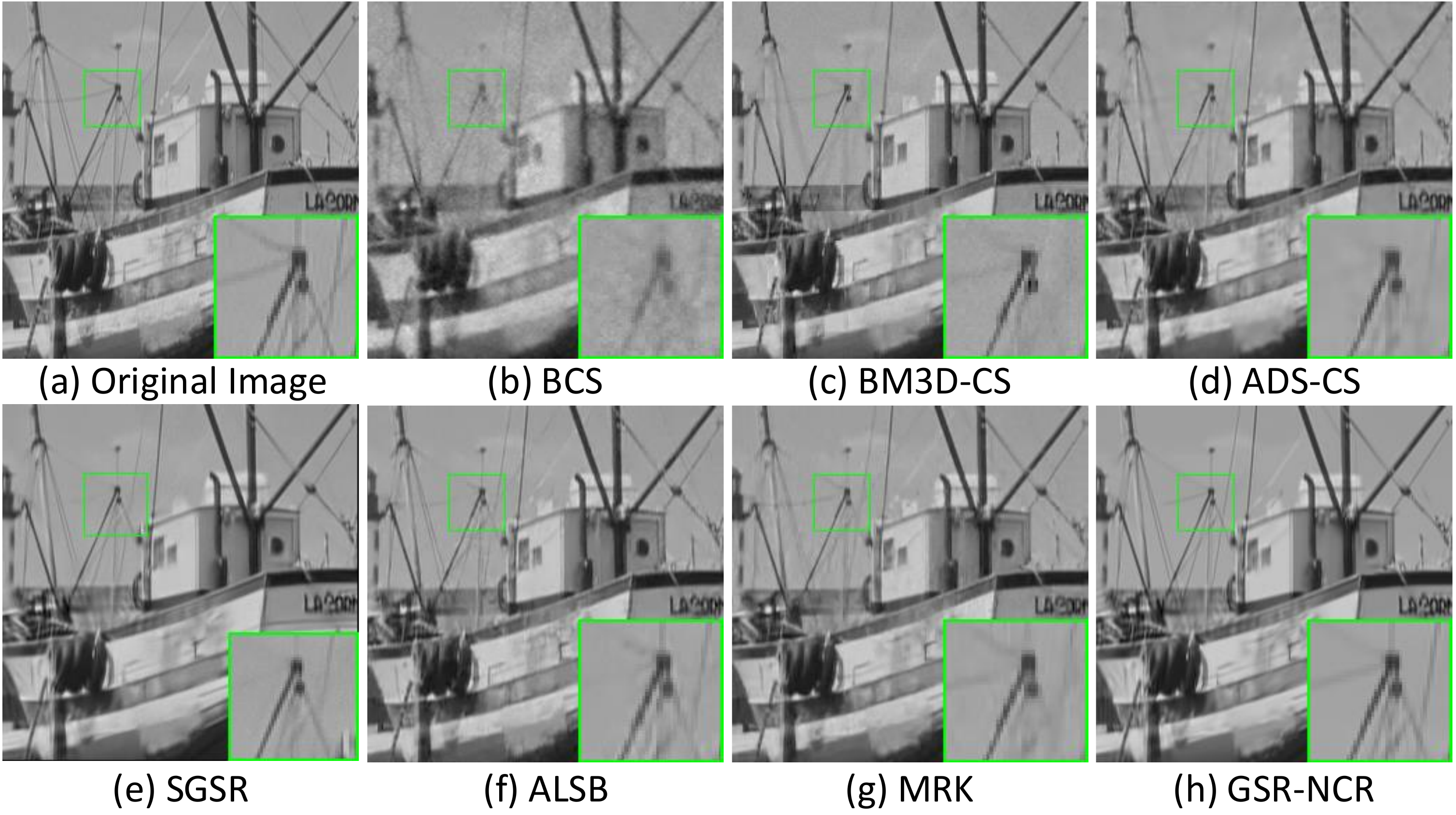}}
\end{minipage}
\caption{CS reconstructed image $\emph{boats}$ with $0.2N$ measurements. (a) Original image; (b) BCS \cite{20} (PSNR=27.05dB, FSIM=0.865); (c) BM3D-CS \cite{21} (PSNR=31.02dB, FSIM=0.931); (d) ADS-CS \cite{12} (PSNR=33.15dB, FSIM=0.951); (e) SGSR \cite{18} (PSNR=32.43dB, FSIM=0.947); (f) ALSB \cite{13} (PSNR= 32.96dB, FSIM=0.951); MRK \cite{42} (PSNR=32.38dB, FSIM=0.948); GSR-NCR (PSNR=\textbf{33.31dB}, FSIM=\textbf{0.953}).}
\label{fig:3}
\end{figure*}

\begin{figure*}[!htbp]
\begin{minipage}[b]{1\linewidth}
  \centering
  \centerline{\includegraphics[width=12cm]{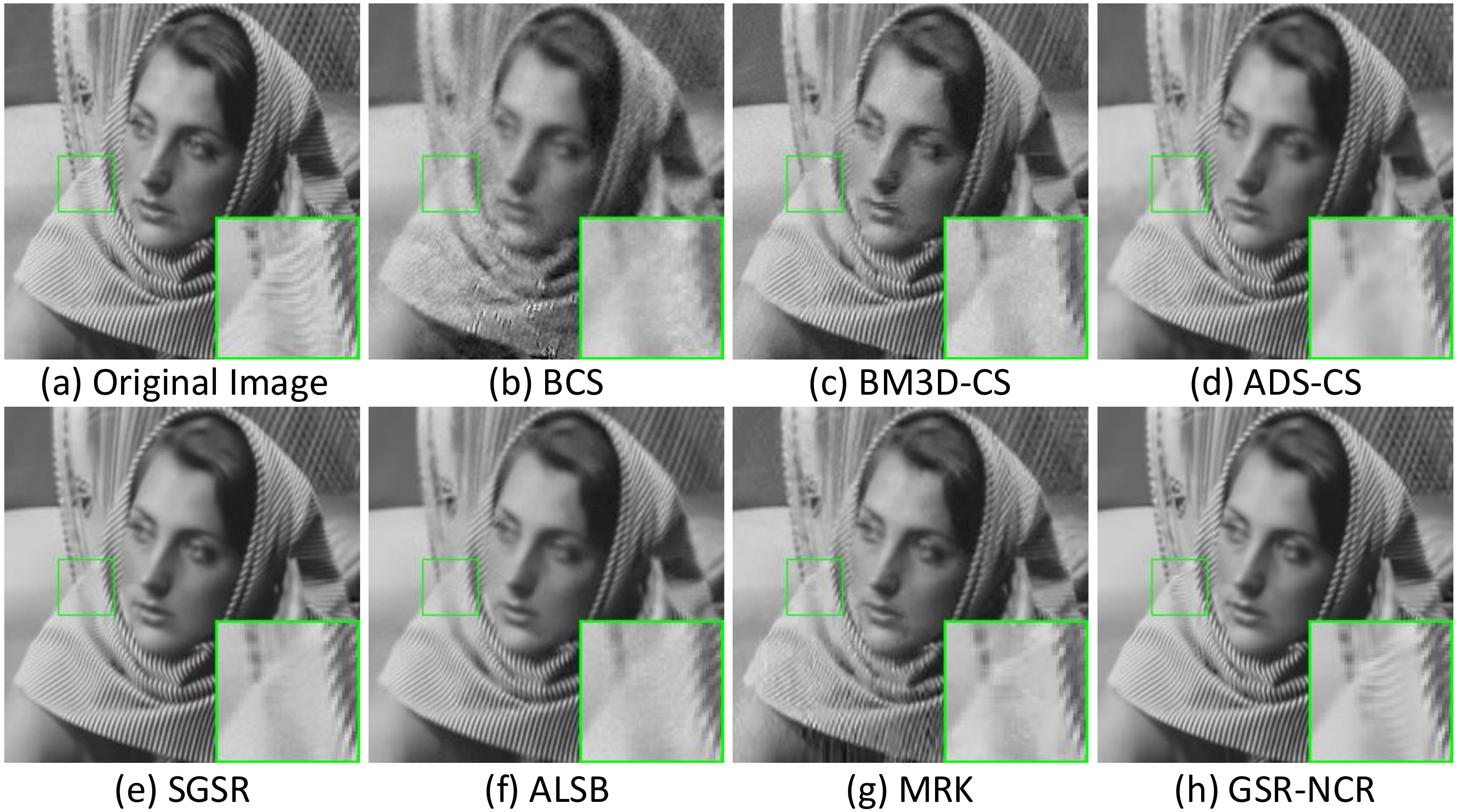}}
\end{minipage}
\caption{CS reconstructed image $\emph{Barbara}$ with $0.2N$ measurements. (a) Original image; (b) BCS \cite{20} (PSNR=22.24dB, FSIM=0.844); (c) BM3D-CS \cite{21} (PSNR=28.82dB, FSIM=0.907); (d) ADS-CS \cite{12} (PSNR=32.27dB, FSIM=0.950); (e) SGSR \cite{18} (PSNR=33.44dB, FSIM=0.962); (f) ALSB \cite{13} (PSNR= 30.72dB, FSIM=0.932); MRK \cite{42} (PSNR=27.99dB, FSIM=0.914); GSR-NCR (PSNR=\textbf{33.93dB}, FSIM=\textbf{0.964}).}
\label{fig:4}
\end{figure*}

\begin{figure*}[!htbp]
\begin{minipage}[b]{1\linewidth}
  \centering
  \centerline{\includegraphics[width=12cm]{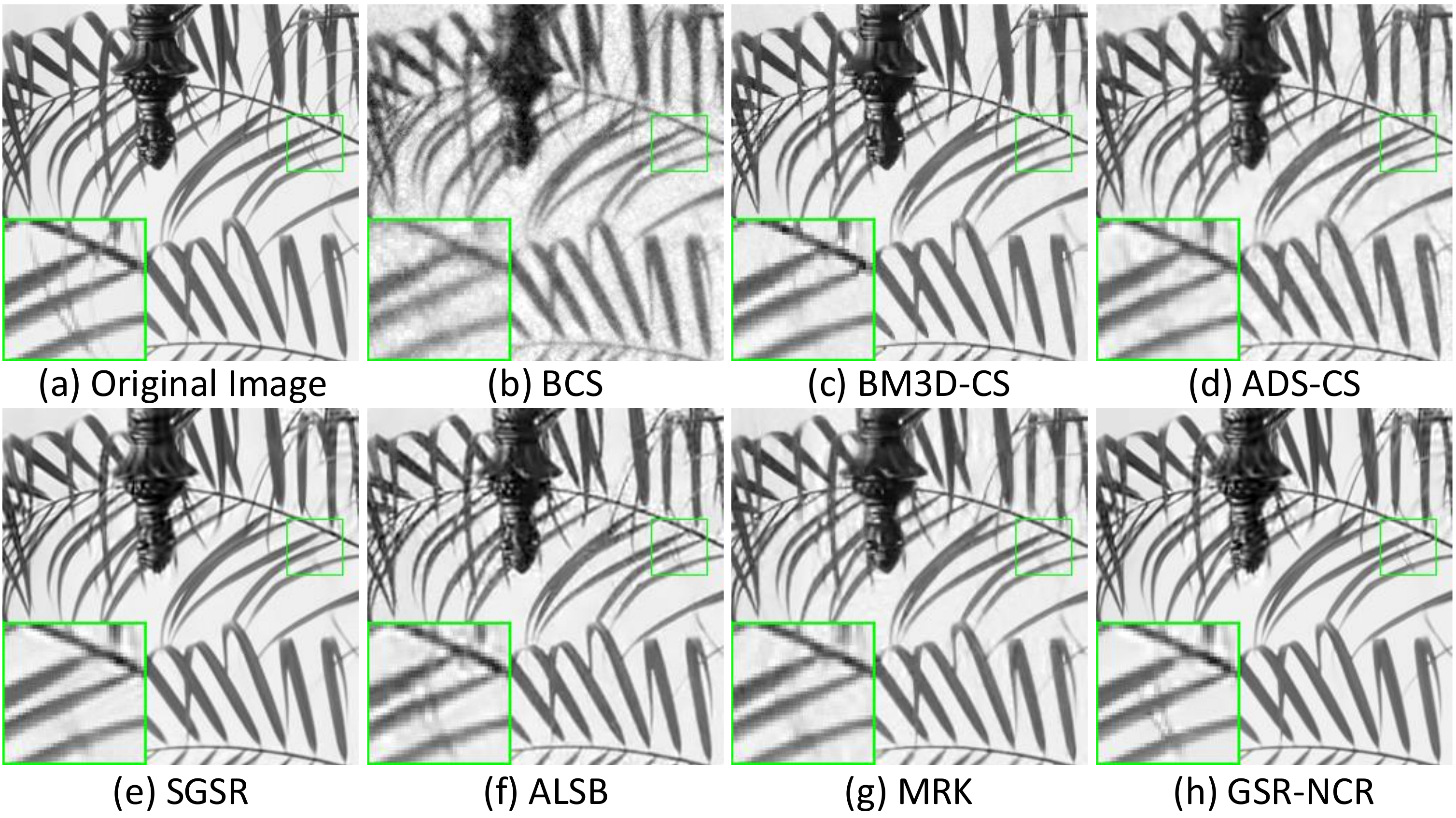}}
\end{minipage}
\caption{CS reconstructed image $\emph{Leaves}$ with $0.2N$ measurements. (a) Original image; (b) BCS \cite{20} (PSNR=21.12dB, FSIM=0.753); (c) BM3D-CS \cite{21} (PSNR=28.13dB, FSIM=0.923); (d) ADS-CS \cite{12} (PSNR=27.88dB, FSIM=0.902); (e) SGSR \cite{18} (PSNR=28.80dB, FSIM=0.938); (f) ALSB \cite{13} (PSNR= 27.15dB, FSIM=0.909); MRK \cite{42} (PSNR=27.75dB, FSIM=0.917); GSR-NCR (PSNR=\textbf{29.03dB}, FSIM=\textbf{0.943}).}
\label{fig:5}
\end{figure*}

\begin{figure*}[!htbp]
\begin{minipage}[b]{1\linewidth}
  \centering
  \centerline{\includegraphics[width=12cm]{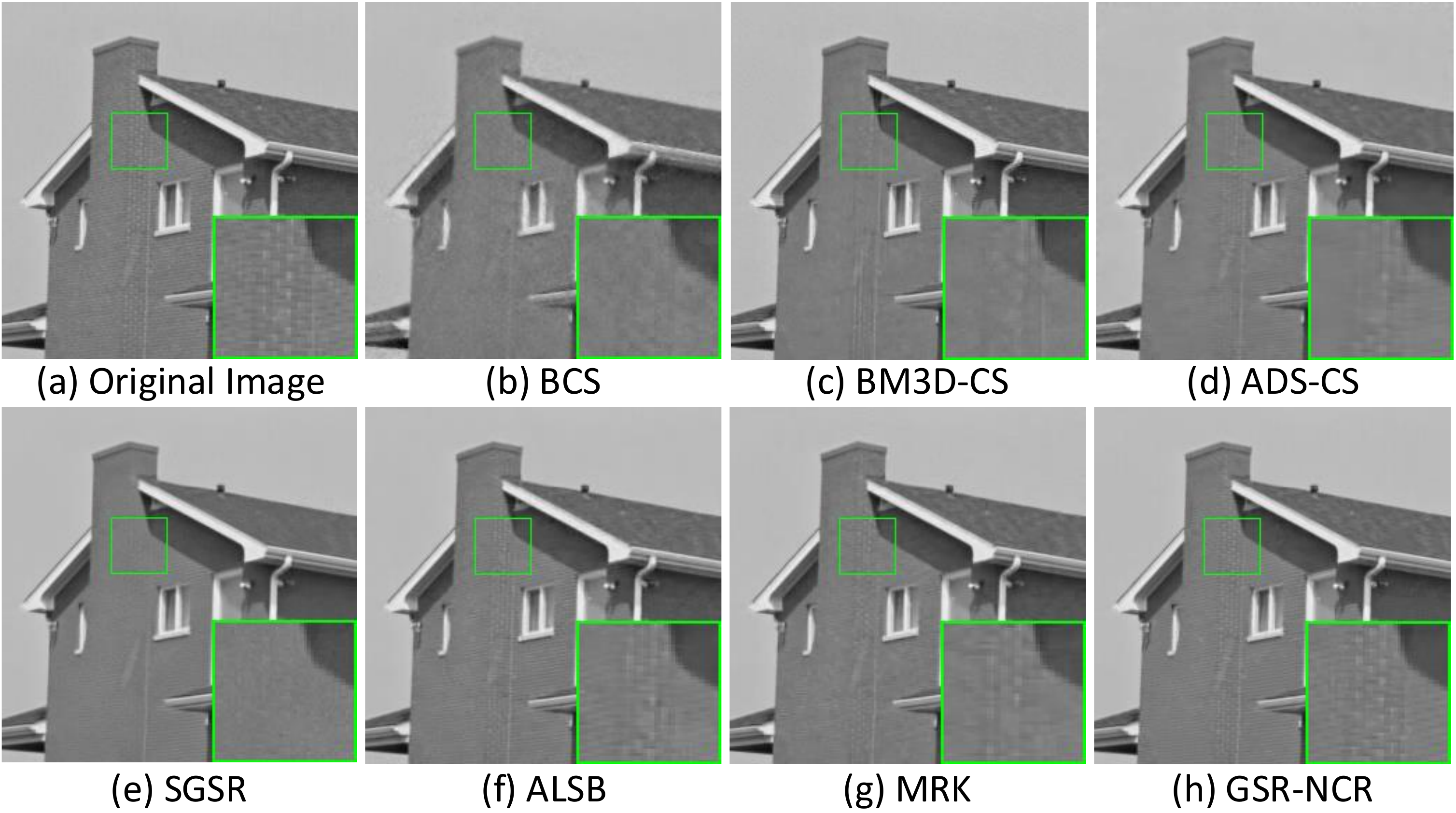}}
\end{minipage}
\caption{CS recovered $\emph{House}$ images with $0.3N$ measurements. (a) Original image; (b) BCS \cite{20} (PSNR=32.85dB, FSIM=0.930); (c) BM3D-CS \cite{21} (PSNR=36.88dB, FSIM=0.969); (d) ADS-CS \cite{12} (PSNR=38.21dB, FSIM=0.967); (e) SGSR \cite{18} (PSNR=37.37dB, FSIM=0.965); (f) ALSB \cite{13} (PSNR= 38.34dB, FSIM=0.973); MRK \cite{42} (PSNR=38.35dB, FSIM=0.973); GSR-NCR (PSNR=\textbf{39.38dB}, FSIM=\textbf{0.980}).}
\label{fig:6}
\end{figure*}

We have compared the proposed GSR-NCR against six other competing approaches including BCS \cite{20}, BM3D-CS \cite{21}, ADS-CS \cite{12}, SGSR \cite{18}, ALSB \cite{13} and MRK \cite{42}. Note that ADS-CS and ALSB are patch-based sparse representation methods for image CS reconstruction. The PSNR and FSIM results by the competing CS reconstruction methods are shown in Table~\ref{lab:2} and Table~\ref{lab:3}, respectively. It can be seen that the proposed GSR-NCR performs competitively compared to other methods. In terms of PSNR, the proposed GSR-NCR achieves 7.89dB, 2.85dB, 1.11dB, 1.20dB, 3.22dB and 2.72dB improvement on average over the BCS, BM3D-CS, ADS-CS, SGSR, ALSB and MRK, respectively. Meanwhile, based on the FSIM, the proposed GSR-NCR achieves 0.1031, 0.0339, 0.0123, 0.0063, 0.0106 and 0.0316 improvement on average over the BCS, BM3D-CS, ADS-CS, SGSR, ALSB and MRK, respectively. The visual comparisons of the image CS reconstruction are shown in Figs.~\ref{fig:3} -~\ref{fig:6}. It can be seen that the BCS, BM3D-CS, ADS-CS, SGSR, ALSB and MRK methods still suffer from some undesirable artifacts or over-smooth phenomena. By contrast, the proposed GSR-NCR not only removes most of the visual artifacts, but also preserves large-scale sharp edges and small-scale fine image details more effectively.
\begin{table*}[!htbp]
\caption{PSNR (dB) Comparisons of BCS \cite{20}, BM3D-CS \cite{21}, ADS-CS \cite{12}, SGSR \cite{18}, ALSB \cite{13}, MRK \cite{42} and the Proposed GSR-NCR.}
\centering  
\small
\begin{tabular}{|c|c|c|c|c|c|c|c|c|c|c|c|}
\hline
\multirow{1}{*}{\textbf{Ratio}} &\multirow{1}{*}{\textbf{Method}}&\multirow{1}{*}{\emph{Barbara}}
&\multirow{1}{*}{\emph{boats}}&\multirow{1}{*}{\emph{Fence}}
&\multirow{1}{*}{\emph{F.print}}&\multirow{1}{*}{\emph{House}}
&\multirow{1}{*}{\emph{Leaves}}&\multirow{1}{*}{\emph{plants}}&\multirow{1}{*}{\emph{straw}}&\multirow{1}{*}{\textbf{\emph{Average}}}\\
 \cline{2-11}
\hline
\multirow{6}{*}{0.2}
   & BCS  & 22.24 & 27.05 & 21.57 & 18.50 & 30.54 & 21.12 & 30.67 & 20.69 & 24.30\\
 \cline{2-11}
   & BM3S-CS  & 28.82 & 31.02 & 26.87 & 19.37 & 35.01 & 28.13 & 34.98 & 20.04 & 28.03\\
 \cline{2-11}
 & ADS-CS     & 32.27 & 33.15 & 28.37 & 22.70 & 35.76 & 27.88 & 35.45 & 23.75 & 29.92\\
 \cline{2-11}
  & SGSR      & 33.44 & 32.43 & \textbf{29.42} & 23.60 & 35.81 & 28.79 & 34.64 & \textbf{24.54} & 30.33\\
 \cline{2-11}
   & ALSB     & 30.72 & 32.96 & 28.41 & \textbf{23.69} & 36.08 & 27.15 & 32.16 & 24.33 & 24.33\\
 \cline{2-11}
 & MRK        & 27.99 & 32.38 & 22.20 & 20.54& 36.36 & 27.75 & \textbf{35.99} & 23.02 & 28.28\\
 \cline{2-11}
  & GSR-NCR  & \textbf{33.93} & \textbf{33.31}& {29.10} & 23.66& \textbf{36.57} & \textbf{29.03} & 35.72 & {24.42} & \textbf{30.72}\\
  \hline
   \cline{2-11}
\hline
\multirow{6}{*}{0.3}
   & BCS    & 25.59 & 28.91 & 23.24 & 19.96 & 32.85 & 23.16 & 32.81& 22.19 & 26.09\\
 \cline{2-11}
 & BM3D-CS  & 33.01 & 34.04 & 30.67 & 23.01 & 36.88 & 32.52 & 38.30 & 22.37 & 31.35\\
 \cline{2-11}
 & ADS-CS   & 35.81 & 36.35& 31.29 & 25.33 & 38.21 & 32.55& 38.45 & 26.58 & 33.07\\
 \cline{2-11}
  & SGSR    & 35.91 & 35.22& 31.56 & 25.84 & 37.37 & 33.00 & 37.20& 27.34 & 32.93\\
 \cline{2-11}
   & ALSB   & 35.00 & 36.42& 30.83 & 25.84 & 38.34 & 31.08& 38.05 & 26.61 & 32.77\\
 \cline{2-11}
 & MRK      & 32.64 & 34.97 & 24.44 & 24.21 & 38.35 & 32.37 & 39.06& 25.52& 31.45\\
 \cline{2-11}
 & GSR-NCR & \textbf{37.19} & \textbf{37.27} & \textbf{32.26}& \textbf{26.35}& \textbf{39.38} & \textbf{34.95} & \textbf{40.10}& \textbf{27.58}& \textbf{34.38}\\
  \cline{2-11}
  \hline
  \multirow{6}{*}{0.4}
   & BCS    & 27.10& 30.56 & 24.81 & 21.67 & 34.65 & 25.07 & 34.77 & 23.71 & 27.79\\
 \cline{2-11}
 & BM3D-CS  & 35.92& 36.69 & 33.84& 25.47 & 38.08 & 35.87 & 41.18 & 24.38& 33.93\\
 \cline{2-11}
  & ADS-CS  & 38.34 & 38.79 & 34.02 & 27.32 & 40.30 & 35.94 & 40.77 & 28.80 & 35.54\\
 \cline{2-11}
  & SGSR    & 37.70& 37.41 & 33.35 & 27.85 & 38.99 & 35.83& 39.23& 29.63 & 35.00\\
 \cline{2-11}
 & ALSB     & 37.19 & 38.92 & 32.83 & 27.70& 40.25 & 34.57 & 40.66 & 28.54 & 35.08\\
 \cline{2-11}
  & MRK     & 36.17& 37.20& 26.63 & 26.83 & 40.04 & 35.53 & 41.64& 27.69 & 33.97\\
 \cline{2-11}
  & GSR-NCR& \textbf{39.23} & \textbf{39.65} & \textbf{34.39} & \textbf{28.53} & \textbf{41.12} & \textbf{38.55} & \textbf{42.48}& \textbf{30.06}& \textbf{36.75}\\
 \cline{2-11}
 \hline
\end{tabular}
\label{lab:2}
\end{table*}
\begin{table*}[!htbp]
\caption{FSIM Comparisons of BCS \cite{20}, BM3D-CS \cite{21}, ADS-CS \cite{12}, SGSR \cite{18}, ALSB \cite{13}, MRK \cite{42} and the Proposed GSR-NCR.}
\centering  
\footnotesize
\begin{tabular}{|c|c|c|c|c|c|c|c|c|c|c|c|}
\hline
\multirow{1}{*}{\textbf{Ratio}} &\multirow{1}{*}{\textbf{Method}}&\multirow{1}{*}{\emph{Barbara}}
&\multirow{1}{*}{\emph{boats}}&\multirow{1}{*}{\emph{Fence}}
&\multirow{1}{*}{\emph{F.print}}&\multirow{1}{*}{\emph{House}}
&\multirow{1}{*}{\emph{Leaves}}&\multirow{1}{*}{\emph{plants}}&\multirow{1}{*}{\emph{straw}}&\multirow{1}{*}{\textbf{\emph{Average}}}\\
 \cline{2-11}
\hline
\multirow{6}{*}{0.2}
   & BCS  & 0.8443 & 0.8654 & 0.7653 & 0.7355 & 0.9011 &0.7531 & 0.8973 &0.7606 & 0.8153\\
 \cline{2-11}
   & BM3S-CS  & 0.9072 & 0.9314 & 0.8325 & 0.8184 & 0.9498 & 0.9231 & 0.9450 & 0.7604 & 0.8835\\
 \cline{2-11}
 & ADS-CS     & 0.9498 & 0.9508 & 0.9181 & 0.8976 & 0.9423 & 0.9015 & 0.9458 & 0.8704 & 0.9220\\
 \cline{2-11}
  & SGSR      & 0.9615 & 0.9468 & \textbf{0.9398} & 0.9208 & 0.9502 & 0.9381 & 0.9431 & \textbf{0.8856} & 0.9357\\
 \cline{2-11}
   & ALSB     & 0.9324 & 0.9514 & 0.9275 & \textbf{0.9226} & {0.9563} & 0.9089 & 0.9145 & 0.8830 & 0.9246\\
 \cline{2-11}
 & MRK        & 0.9135 & 0.9476 & 0.7765 & 0.8397 & \textbf{0.9586} & 0.9169 & \textbf{0.9555} & 0.8418 & 0.8938\\
 \cline{2-11}
  & GSR-NCR  &\textbf{0.9642} & \textbf{0.9526} & 0.9377 &0.9224 & 0.9508 &\textbf{0.9431} &0.9505 & {0.8852} & \textbf{0.9383}\\
  \hline
   \cline{2-11}
\hline
\multirow{6}{*}{0.3}
   & BCS    & 0.8782 & 0.8997 & 0.8345 & 0.8149 & 0.9299 & 0.8018 & 0.9276 & 0.8266 &0.8641\\
 \cline{2-11}
 & BM3D-CS  & 0.9587 & 0.9630 & 0.9572 & 0.9111 & 0.9690 & 0.9601 &0.9714 & 0.8322 & 0.9403\\
 \cline{2-11}
 & ADS-CS   & 0.9733 & 0.9728 & 0.9521 & 0.9408 & 0.9667 & 0.9550 & 0.9697 & 0.9220 & 0.9565\\
 \cline{2-11}
  & SGSR    & 0.9762 & 0.9684 & 0.9600 & 0.9482 & 0.9648 & 0.9676 & 0.9654 & 0.9316 & 0.9603\\
 \cline{2-11}
   & ALSB   & 0.9729 & 0.9746 & 0.9557 & 0.9475 & 0.9732 & 0.9511 & 0.9736 & 0.9226 & 0.9589\\
 \cline{2-11}
 & MRK      & 0.9611 & 0.9687 & 0.8415 & 0.9225 & 0.9727 & 0.9598 & 0.9768 & 0.9040 & 0.9384\\
 \cline{2-11}
 & GSR-NCR &\textbf{0.9816} & \textbf{0.9783} & \textbf{0.9664} &\textbf{0.9534} & \textbf{0.9795} &\textbf{0.9799} & \textbf{0.9819} & \textbf{0.9351} & \textbf{0.9695}\\
  \cline{2-11}
  \hline
  \multirow{6}{*}{0.4}
   & BCS    & 0.9068 & 0.9248 & 0.8807 & 0.8747 & 0.9490 & 0.8422 & 0.9479 & 0.8748 & 0.9001\\
 \cline{2-11}
 & BM3D-CS  & 0.9777 & 0.9805 & 0.9758 & 0.9452 & 0.9781 & 0.9803 & 0.9855 & 0.8848 & 0.9635\\
 \cline{2-11}
  & ADS-CS  & 0.9837 & 0.9835 & 0.9726 & 0.9608 & 0.9803 & 0.9763 & 0.9816 & 0.9487 & 0.9734\\
 \cline{2-11}
  & SGSR    & 0.9836 & 0.9793 & 0.9728 & 0.9653 & 0.9759 & 0.9799 & 0.9777 & 0.9570 & 0.9739\\
 \cline{2-11}
 & ALSB     & 0.9827 & 0.9840 & 0.9702 & 0.9638 & 0.9824 & 0.9738 & 0.9840 & 0.9479 & 0.9736\\
 \cline{2-11}
  & MRK     & 0.9795 & 0.9802 & 0.8979 & 0.9539 & 0.9819 & 0.9783 & 0.9873 & 0.9377 & 0.9620\\
 \cline{2-11}
  & GSR-NCR& \textbf{0.9879} & \textbf{0.9867} &\textbf{0.9784} & \textbf{0.9702} & \textbf{0.9862} & \textbf{0.9894} &\textbf{0.9892} & \textbf{0.9609} & \textbf{0.9811}\\
 \cline{2-11}
 \hline
\end{tabular}
\label{lab:3}
\end{table*}

\subsection{Effect of the number of the best matched patches}
\begin{figure*}[!htbp]
\begin{minipage}[b]{1\linewidth}
  \centerline{\includegraphics[width=12cm]{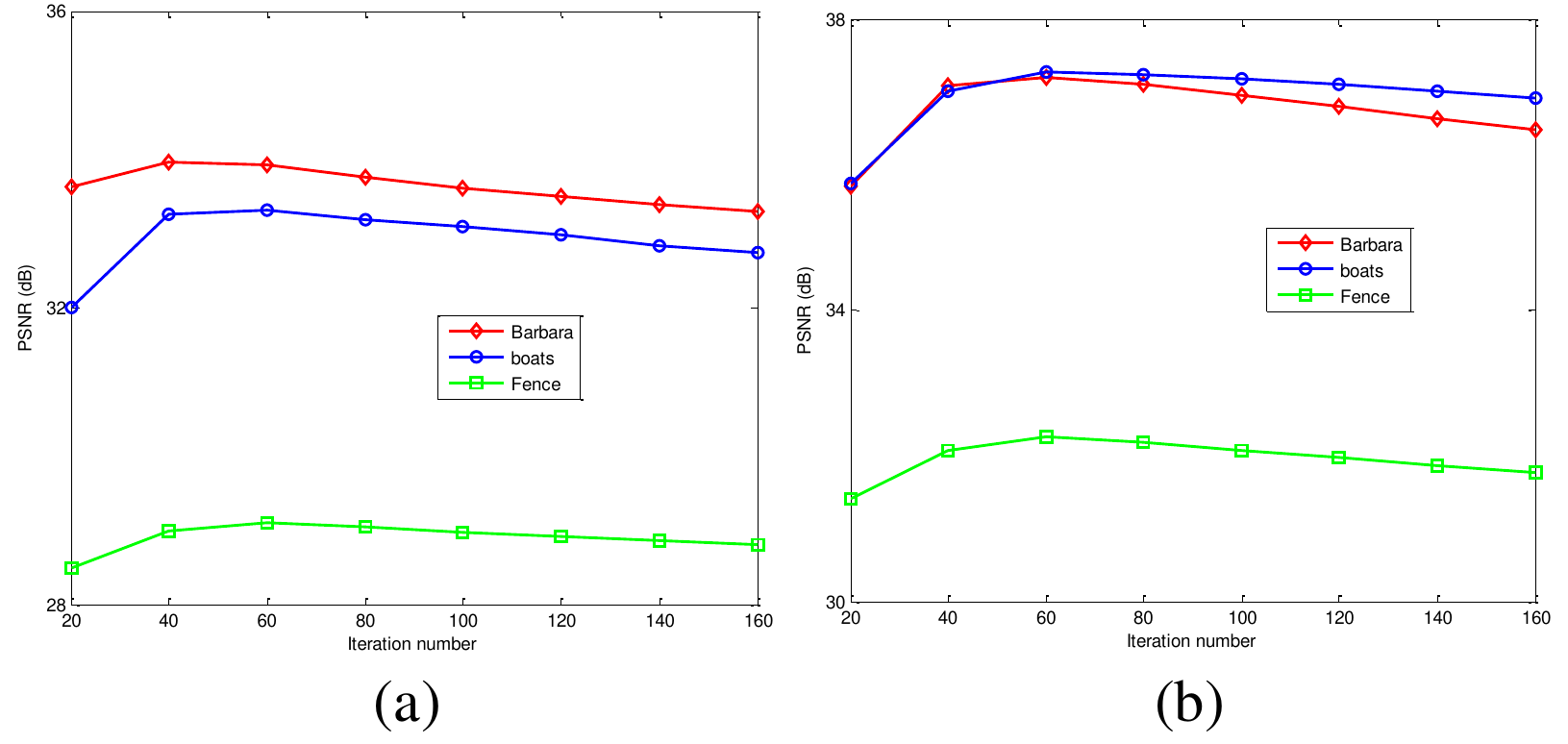}}
\end{minipage}
\caption{Performance comparison with different matched patch numbers $c$ for image CS reconstruction . (a) PSNR results achieved by different $c$ in the case of 0.2$N$ measurements. (b)  PSNR results achieved by different $c$ in the case of 0.3$N$ measurements.}
\label{fig:7}
\end{figure*}

We have discussed how to select the best matching patch numbers $c$ for the performance of the proposed GSR-NCR. Specifically, to investigate the sensitivity of our method against $c$, two experiments were conducted with respect to different $c$, ranging from 20 to 160, in the case of 0.2$N$ and $0.3N$ measurements, respectively. The results with different $c$ are shown in Fig.~\ref{fig:7}. It can be seen  that all the curves are  almost flat, showing the performance of the proposed GSR-NCR scheme is insensitive to $c$. The best performance of each case was usually achieved with $c$ in the range [40,80]. Therefore, in this paper $c$ was empirically set to be 60.
\begin{figure*}[!htbp]
\begin{minipage}[b]{1\linewidth}
  \centering
  \centerline{\includegraphics[width=12cm]{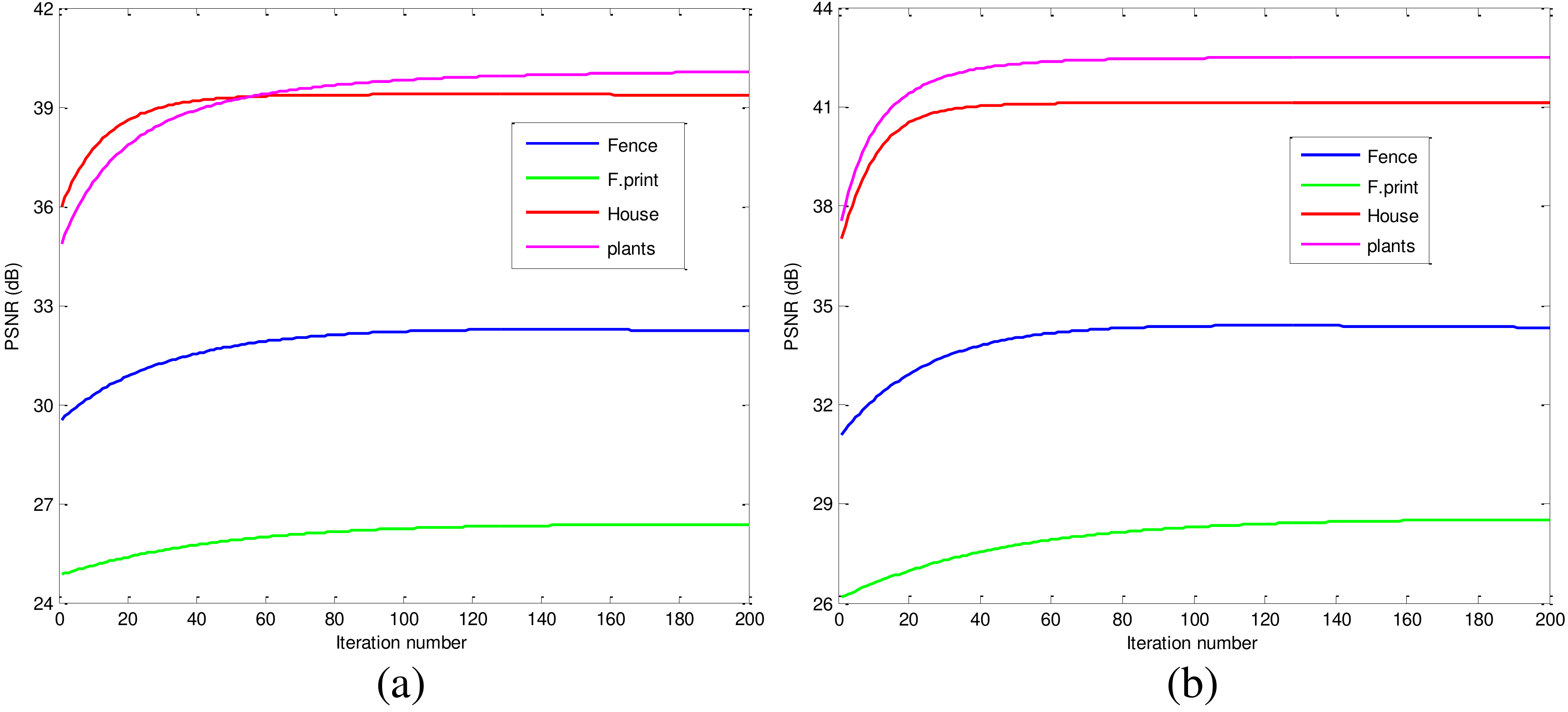}}
\end{minipage}
\caption{Convergence analysis of the proposed GSR-NCR. (a) PSNR results versus iteration numbers for image CS reconstruction with 0.3$N$ measurements; (b) PSNR results versus iteration numbers for image CS reconstruction with 0.4$N$ measurements.}
\label{fig:8}
\end{figure*}
\subsection{Convergence analysis}
Since the proposed GSR-NCR model (Eq.~\eqref{eq:6}) is non-convex, it is difficult to give its theoretical proof for global convergence. Here, we only provide empirical evidence to illustrate the good convergence of the proposed CS reconstruction method. Fig.~\ref{fig:8} illustrates the convergent performance of the proposed GSR-NCR. It shows the curves of the PSNR values versus the iteration numbers for four test images with $0.3N$ and $0.4N$ measurements, respectively. One can observe that with the increase of the iteration numbers, the PSNR curves gradually increase and ultimately become flat and stable, showing good stability of the proposed non-convex GSR-NCR model.

\section{Conclusion}
\label{5}
In this paper, we proposed a efficient method for image CS reconstruction using group-based sparse representation model, which is able to more accurately enforce the local sparsity and nonlocal self-similarity of images simultaneously in a unified framework. Different from the typical sparsity-promoting convex $\ell_1$ minimization methods, we extend the non-convex weighted $\ell_p$ ($0<p<1$) penalty function on group sparse coefficients of the data matrix to replace the convex $\ell_1$-norm. To reduce the computational complexity, we learn the principle component analysis (PCA) based dictionary for each group to substitute for the dictionary with a high computational complexity learned from natural image dataset.  Furthermore, to make the proposed model tractable and robust, an efficient iterative shrinkage/thresholding algorithm was adopted to solve the non-convex minimization problem. Experimental results have shown that the proposed method not only outperforms many state-of-the-art methods both quantitatively and qualitatively, but also results in fine stability.

\end{document}